\documentclass[nologo,url,11pt,a4paper]{ETHpaper}
\pdfoutput=1

\usepackage{amsmath, amsthm, amssymb}

\DeclareMathAlphabet{\mathitbf}{OML}{cmm}{b}{it}
\DeclareMathAlphabet{\mathbf}{OML}{cmm}{b}{it}
\usepackage{graphicx}                  
\usepackage{subfig}                    
\usepackage[numbers,sort&compress]{natbib}     

\usepackage{amsmath}
\usepackage{amssymb}

\usepackage{color} 

\title{Positive words carry less information than negative words}

\author{David Garcia, Antonios Garas, and Frank Schweitzer}

\address{Chair of Systems Design, ETH Zurich, Kreuzplatz 5, 8032 Zurich,
  Switzerland}

\authoralternative{D. Garcia, A. Garas, F. Schweitzer}

\reference{\textsl{EPJ Data Science} \textbf{1} 3 (2012)}

\www{\url{http://www.sg.ethz.ch}}

\begin{document}
\maketitle
\renewcommand{\thefootnote}{ \fnsymbol{footnote}} 

\begin{abstract}
  We show that the frequency of word use is not only determined by the
  word length \cite{Zipf1935} and the average information content
  \cite{Piantadosi2011}, but also by its emotional content. We have
  analyzed three established lexica of affective word usage in English,
  German, and Spanish, to verify that these lexica have a neutral,
  unbiased, emotional content.  Taking into account the frequency of word
  usage, we find that words with a positive emotional content are more
  frequently used. This lends support to Pollyanna hypothesis
  \cite{Boucher1969} that there should be a positive bias in human
  expression.  We also find that negative words contain more information
  than positive words, as the informativeness of a word increases
  uniformly with its valence decrease. Our findings support earlier
  conjectures about (i) the relation between word frequency and
  information content, and (ii) the impact of positive emotions on
  communication and social links.
\end{abstract}

\newcommand{\mean}[1]{\left\langle #1 \right\rangle}
\newcommand{\abs}[1]{\left| #1 \right|}


\begin{center}
\end{center}

\section{Introduction}

One would argue that human languages, in order to facilitate social
relations, should be biased towards positive emotions. This question
becomes particularly relevant for sentiment classification, as many
tools assume as null hypothesis that human expression has neutral
emotional content \cite{Pang2008,Thelwall2011a}, or reweight positive
and negative emotions \cite{Taboada2011} without a quantification of
the positive bias of emotional expression. We have tested and measured
this bias in the context of online written communication by analyzing
three established lexica of affective word usage. These lexica cover
three of the most used used languages on the Internet, namely English
\cite{Bradley1999}, German \cite{Vo2009}, and Spanish
\cite{Redondo2007}. The emotional content averaged over all the words
in each of them is neutral. Considering, however, the everyday usage
frequency of these words we find that the overall emotion of the three
languages is strongly biased towards positive values, because words
associated with a positive emotion are more frequently used than those
associated with a negative emotion.

Historically, the \emph{frequency} of words was first analyzed by Zipf
\cite{Zipf1935, Zipf1949} showing that frequency predicts the
\emph{length} of a word as result of a principle of least effort.
Zipf's law highlighted fundamental principles of organization in human
language \cite{FerreriCancho2003}, and called for an interdisciplinary
approach to understand its origin
\cite{Hauser2002,Kosmidis2006,Havlin1995} and its relation to word
meaning \cite{Piantadosi2011b}.  Recently Piantadosi et
al.~\cite{Piantadosi2011} extended Zipf's approach by showing that, in
order to have efficient communication, word length increases with
information content.  Further discussions \cite{Griffiths2011,
  Reilly2011, Piantadosi2011b} highlighted the relevance of
\emph{meaning} as part of the communication process as, for example,
more abstract ideas are expressed through longer
words~\cite{Reilly2007}.  Our work focuses on one particular aspect of
meaning, namely the \emph{emotion} expressed in a word, and how this
is related to word frequency and information content.  This approach
requires additional data beyond word length and frequency, which
became available thanks to large datasets of human behaviour on the
Internet. Millions of individuals write text online, for which a
quantitative analysis can provide new insights into the structure of
human language and even provide a validation of social theories
\cite{Lazer2009}.  Sentiment analysis techniques allow to quantify the
emotions expressed through posts and messages \cite{Thelwall2011a,
  Taboada2011}.  Recent studies have provided statistical analyses
\cite{Bollen2010, Golder2011, Chmiel2011, Dodds2011} and modelling
approaches \cite{Schweitzer2010, Garcia2011} of individual and
collective emotions on the Internet.

An emotional bias in written expressions, however, would have a strong
impact, as it shifts the balance between positive and negative
expressions. Thus, for all researchers dealing with emotions in
written text it would be of particular importance to know about such
bias, how it can be quantified, and how it affects the baseline, or
reference point, for expressed emotions.  Our investigation is devoted
to this problem by combining two analyses, (i) quantifying the
emotional content of words in terms of valence, and (ii) quantifying
the frequency of word usage in the whole indexable web
\cite{Brants2009}.  We provide a study of the baseline of written
emotional expression on the Internet in three languages that span more
than 67.7\% of the websites \cite{Wikipedia2011}: English (56.6\%),
German (6.5\%), and Spanish (4.6\%). These languages are used everyday
by more than 805 million users, who create the majority of the content
available on the Internet.

In order to link the emotionality of each word with the information it
carries, we build on the recent work of Piantadosi et
al. \cite{Piantadosi2011}. This way, we reveal the importance of
emotional content in human communication which influences the information
carried by words. While the rational process that optimizes communication
determines word lengths by the information they carry, we find that the
emotional content affects the word frequency such that positive words
appear more frequently. This points towards an emotional bias in used
language and supports Pollyanna hypothesis \cite{Boucher1969}, which
asserts that there is a bias towards the usage of positive
words. Furthermore, we extend the analysis of information content by
taking into account word context rather than just word frequency. This
leads to the conclusion that positive words carry less information than
negative ones. In other words, the informativeness of words highly
depends on their emotional polarity.

We wish to emphasize that our work distinguishes itself both regarding
its methodology and its findings from a recent article
\cite{Kloumann2012}. There, the authors claim a bias in the amount of
positive versus negative words in English, while no relation between
emotionality and frequency of use was found. A critical examination of
the conditions of that study shows that the quantification of emotions
was done in an uncontrolled setup through the Amazon Mechanical Turk.
Participants were shown a scale similar to the ones used in previous
works \cite{Bradley1999,Vo2009,Redondo2007}, as explained in
\cite{Dodds2011}. Thanks to the popular usage of the Mechanical Turk, the
authors evaluated more than $10.000$ terms from the higher frequency
range in four different corpora of English expression. However, the
authors did not report any selection criterion for the participant
reports, opposed to the methodology presented in \cite{Bohannon2011}
where up to 50\% of the participants had to be discarded in some
experiments.

Because of this lack of control in their experimental setup, the positive
bias found in \cite{Kloumann2012} could be easily explained as an
acquiescent bias \cite{Knowles1997, Bentler1969}, a result of the human
tendency to agree in absence of further knowledge or relevance.  In
particular, this bias has been repeatedly shown to exist in self
assessments of emotions \cite{Russell1979, Yik2011}, requiring careful
response formats, scales, and analyses to control for it. Additionally,
the wording used to quantify word emotions in \cite{Kloumann2012}
(\textit{happiness}), could imply two further methodological biases: The
first one is a possible social desirability bias \cite{Robinson1991}, as
participants tend to modify their answers towards more socially
acceptable answers.  The positive social perception of displaying
happiness can influence the answers given by the participants of the
study.  Second, the choice of the word \textit{happiness} implies a
difference compared with the standard psychological term \textit{valence}
\cite{Russell1980}. Valence is interpreted as a static property of the
word while happiness is understood as a dynamic property of the surveyed
person when exposed to the word. This kind of framing effects have been
shown to have a very large influence in survey results. For example, a
recent study \cite{Bryan2011} showed a large change in the answers by
simply changing \textit{voting} for \textit{being a voter} in a voter
turnout survey.

Hence, there is a strong sensitivity to such influences which are not
controlled for in \cite{Kloumann2012}. Because of all these limitations,
in our analysis we chose to use the current standard lexica of word
valence. These lexica, albeit limited to 1000 to 3000 words, were
produced in three controlled, independent setups, and provide the most
reliable estimation of word emotionality for our analysis. Our results on
these lexica are consistent with recent works on the relation between
emotion and word frequency \cite{Augustine2011, Rozin2010} for English in
corpora of limited size.

\section{Results}

\subsection{Frequency of emotional words}
\begin{figure*}[ht!]
 \centering
  \includegraphics[width=0.95\textwidth]{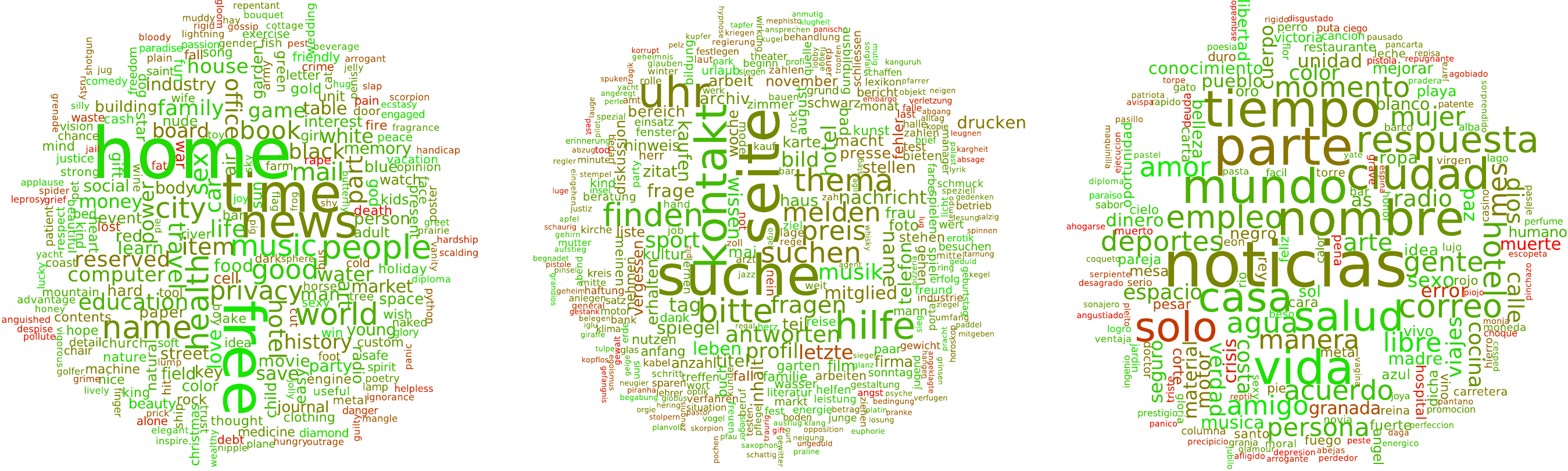}
  \caption{ {\bf Emotion word clouds with frequencies calculated from
      Google's crawl.} In each word cloud for English (left), German
    (middle), and Spanish (right), the size of a word is proportional
    to its frequency of appearance in the trillion-token Google N-gram
    dataset \cite{Brants2009}. Word colors are chosen from red
    (negative) to green (positive) in the valence range from
    psychology studies\cite{Bradley1999,Vo2009,Redondo2007}.  For the
    three languages, positive words predominate on the Internet.}
 \label{fig:GGS-0}
\end{figure*}

In detail, we have analyzed three lexica of affective word usage which
contain 1034 English words, 2902 German words and 1034 Spanish words,
together with their emotional scores obtained from extensive human
ratings. These lexica have effectively established the standard for
emotion analyses of human texts \cite{Dodds2009}.  Each word in these
lexica is assigned a set of values measuring different aspects of word
emotionality. The three independent studies that generated the lexica
for English \cite{Bradley1999}, German \cite{Vo2009}, and Spanish
\cite{Redondo2007} used the Self-Assessment Mannequin (SAM) method to
ask participants about the different emotion values associated to each word
in the lexicon. One of these values, a scalar variable $v$ called
valence, represents the degree of pleasure induced by the emotion
associated with the word, and it is known to explain most of the
variance in emotional meaning\cite{Russell1980}. In this article, we use
$v$ to quantify word emotionality.

\begin{figure}[ht!]
 \centering
  \includegraphics[width=0.47\textwidth]{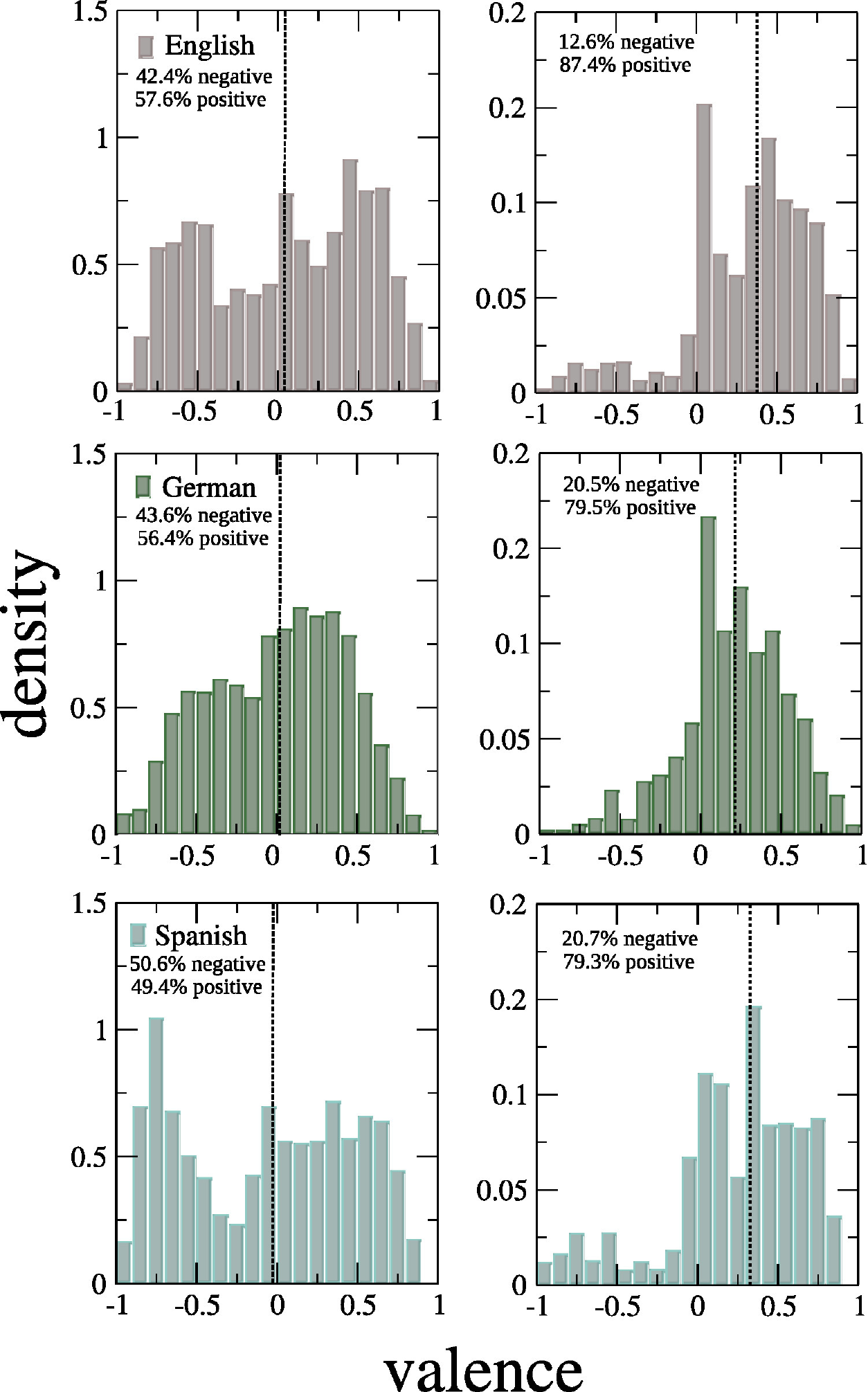}
  \caption{ 
    {\bf Distributions of word emotions weighted by the
      frequency of word usage.} (left panel) Distributions of reported
    valence values for words in English (top panel, lexicon:
    \cite{Bradley1999}, 1034 entries), German (middle panel, lexicon:
    \cite{Vo2009}, 2902 entries), and Spanish (bottom panel, lexicon:
    \cite{Redondo2007}, 1034 entries), normalized by the size of the
    lexica.  Average valence (median) 0.048 (0.095) for English, 0.021
    (0.067) for German, and -0.065 (-0.006) for Spanish.  (right
    panel) Normalized distributions of reported valence values
    weighted by the frequency of word usage, obtained from the same
    lexica.  Average valence (median) 0.314 (0.375) for English, 0.200
    (0.216) for German, and 0.238 (0.325) for Spanish. The dashed
    lines indicate the median. Inset numbers: ratio of positive and
    negative areas in the corresponding distributions.}
 \label{fig:GGS-1}
\end{figure}

In each lexicon, words were chosen such that they evenly span the full
range of valence.\footnote{The lexica focus on single words rather
  than on phrases or longer expressions.}  In order to compare the
emotional content of the three different languages, we have rescaled
all values of $v$ to the interval [-1,1].  As shown in the left panel
of Fig. \ref{fig:GGS-1}, indeed, the average valence, as well as the
median, of all three lexica is very close to zero, i.e. they do not
provide an emotional bias. This analysis, however, neglects the actual
frequency of word usage, which is highly skew distributed
\cite{Zipf1935, Zipf1949}.  For our frequency estimations we have used
Google's $N$-gram dataset \cite{Brants2009} which, with $10^{12}$
tokens, is one of the largest datasets available about real human text
expressions on the Internet.  For our analysis, we have studied the
frequency of the words which have an affective classification in the
respective lexicon in either English, German, or
Spanish. Fig. \ref{fig:GGS-0} shows emotion word clouds for the three
languages, where each word appears with a size proportional to its
frequency. The color of a word is chosen according to its valence,
ranging from red for $v=-1$ to green for $v=+1$. It is clear that
green dominates over red in the three cases, as positive emotions
predominate on the Internet. Some outliers, like ``home'', have a
special higher frequency of appearance in websites, but as we show
later, our results are consistent with frequencies measured from
traditional written texts like books.

In a general setup, the different usage of words with the same valence
is quite obvious.  For example, both words ``party'' and ``sunrise''
have the same positive valence of 0.715, however the frequency of
``party'' is 144.7 per one million words compared to 6.8 for
``sunrise''. Similarly, both ``dead'' and ``distressed'' have a
negative valence of -0.765, but the former appears 48.4 times per one
million words, the latter only 1.6 times. Taking into account all
frequencies of word usage, we find for all three languages that the
median shifts considerably towards positive values. This is shown in
the right panel of Fig.~\ref{fig:GGS-1}.  Wilcoxon tests show that the
means of these distributions are indeed different, with an estimated
difference in a 95\% confidence interval of $0.257 \pm 0.032$ for
English, $0.167 \pm 0.017$ for German, and $0.287 \pm 0.035$ for
Spanish. Hence, with respect to usage we find evidence that the
language used on the Internet is emotionally charged, i.e. significantly
different from being neutral. This affects quantitative analyses of
the emotions in written text, because the ``emotional reference
point'' is not at zero, but at considerably higher valence values
(about 0.3).

\subsection{Relation between information and valence}

Our analysis suggests that there is a definite relation between word
valence and frequency of use. Here we study the role of emotions in
the communication process building on the relation between information
measures and valence. While we are unable to measure information
perfectly, we can approximate its value given the frequencies of words
and word sequences. First we discuss the relation between word valence
and information content estimated from the simple word occurrences,
namely self-information. Then we explain how this extends when the
information is measured taking into account the different contexts in
which a word can appear. The self-information of a word, $I(w)$
\cite{Cover1991} is an estimation of the information content from its
probability of appearance, $P(w)$, as 
\begin{equation}
  I(w) = -{\log}P(w)
\label{eq:selfinfo}
\end{equation}
Frequency-based information content metrics like self-information are
commonly used in computational linguistics to systematically analyze
communication processes. Information content is a better predictor for
word length than word frequency \cite{Piantadosi2011, Jaeger2010}, and
the relation between information content and meaning, including
emotional content, is claimed to be crucial for the way humans
communicate \cite{Griffiths2011, Reilly2011, Piantadosi2011b}. We use
the self-information of a word as an estimation of information content
for a context size of 1, to build up later on larger context
sizes. This way, we frame our analysis inside the larger framework of
N-gram information measures, aiming at an extensible approach that can
be incorporated in the fields of computational linguistics and
sentiment analysis.

For the three lexica, we calculated $I(w)$ of each word and linked it
to its valence, $v(w)$. As defined in eq. \ref{eq:selfinfo}, very
common words provide less information than very unusual ones, but this
nonlinear mapping between frequency and self-information makes the
latter more closely related to word valence than the former.  The
first two lines of Table \ref{tab:GGS-2} show the Pearson's correlation
coefficient of word valence and frequency $\rho(v, f)$, followed by
the correlation coefficient between word valence and self-information,
$\rho(v, I)$. For all three languages, the absolute value of the
correlation coefficient with $I$ is larger than with $f$, showing that
self-information provides more knowledge about word valence than plain
frequency of use.

\begin{figure*}[ht]
 \centering
  \includegraphics[width=0.65\textwidth]{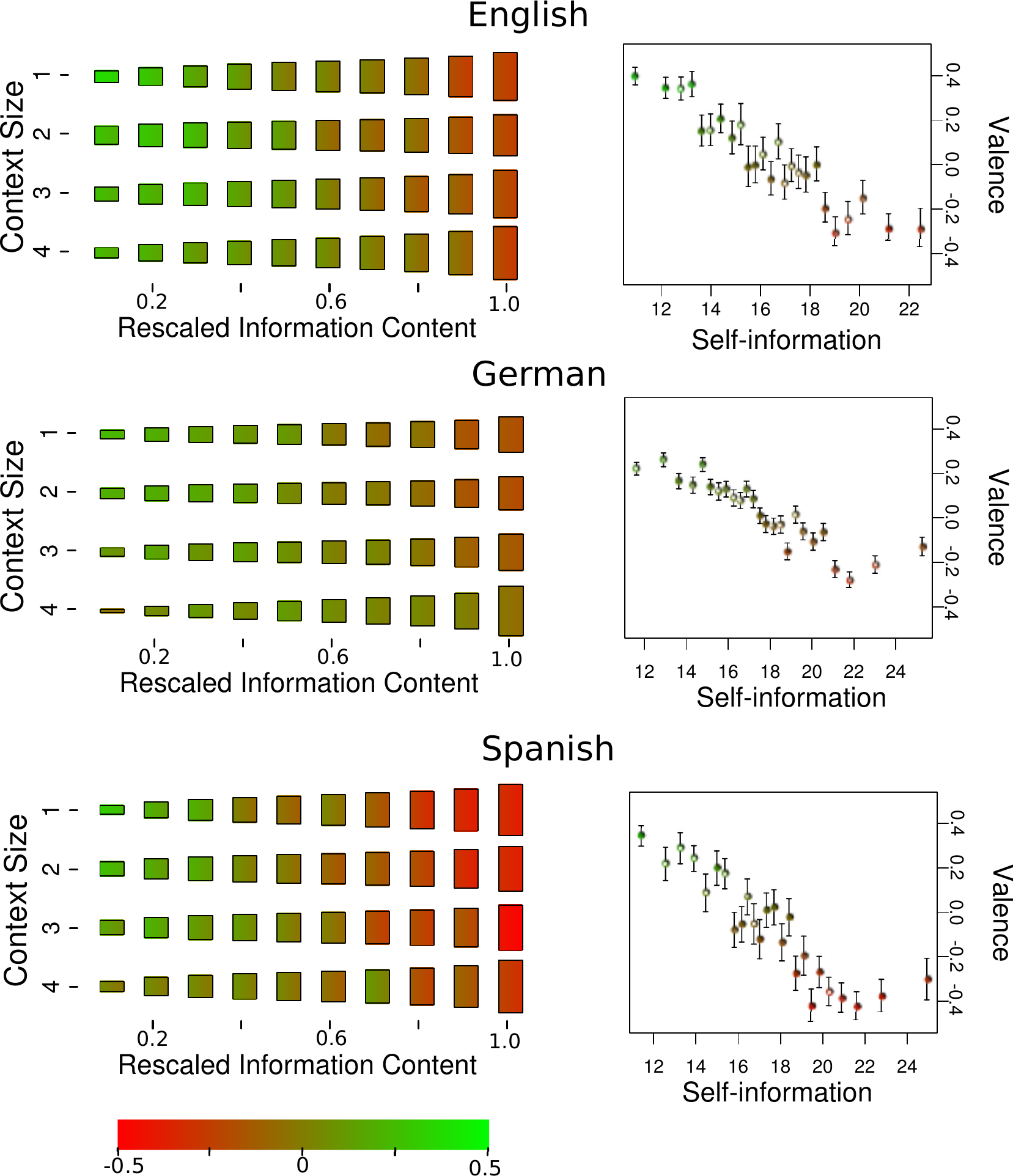}
  \caption{{\bf Relation between information measures and valence.}
    Each graphic on the left column shows the relation between valence
    and information content measured up to a context of size four. Each
    bar represents a bin containing 10\% of the words in the lexicon,
    with a size proportional to the average information content of the
    words in the bar.  The color of each bar ranges from red to green,
    representing the average valence of the words in the corresponding
    bin. Each bar has a color gradient according to the standard error
    of the valence mean. Information content has been rescaled so it
    can be compared among context sizes. For all three languages and
    context sizes, negativity increases with information content. The
    second column shows the relation between word self-information and
    valence for English, German, and Spanish.  Average valence is
    shown for bins that contain 5\% of the data, with error bars
    showing the standard error. For all the three languages, valence
    clearly decreases with the self-information of the word,
    i.e. positive words carry less information than negative words.}
 \label{fig:GGS-2}
\end{figure*}

The right column of Fig. \ref{fig:GGS-2} shows in detail the relation
between $v$ and $I$. From the clear negative correlation found for all
three languages (between -0.3 and -0.4), we deduce that words with
less information content carry more positive emotions, as the average
valence decreases along the self-information range. As mentioned
before the Pearson's correlation coefficient between word valence and
self-information, $\rho(v, I)$, is significant and negative for the
three languages (Table \ref{tab:GGS-2}). Our results outperform a
recent finding ~\cite{Augustine2011} that, while focusing on
individual text production, reported a weaker correlation (below 0.3)
between the logarithm of word usage frequency and valence. This
previous analysis was based on a much smaller data set from Internet
discussions (in the order of $10^{8}$ tokens) and the same English
lexicon of affective word usage \cite{Bradley1999} we used. Using a
much higher accuracy in estimating word frequencies and extending the
analysis to three different languages, we were able to verify that
there is a significant relation between the emotional content of a
word and its self-information, impacting the frequency of usage.

\begin{table}[ht]
  \begin{center}
    \begin{tabular*}{0.8\linewidth}{@{\extracolsep{\fill}}llll}
      \hline
      & English & German & Spanish \cr \hline
      $\rho(v, f)$ & 0.222 $^{**}$ & 0.144 $^{**}$ & 0.236 $^{**}$\cr
      $\rho(v, I)$ &-0.368 $^{**}$ &-0.325 $^{**}$ & -0.402 $^{**}$\cr
      $\rho(v, I')$ & -0.294 $^{**}$ & -0.222 $^{**}$ & -0.311 $^{**}$ \cr 
      $\rho(v, I_2)$ & -0.332 $^{**}$ &  -0.301  $^{**}$ & -0.359 $^{**}$ \cr
      $\rho(v, I_3)$ & -0.313  $^{**}$ & -0.201  $^{**}$ & -0.340  $^{**}$ \cr
      $\rho(v, I_4)$ & -0.254  $^{**}$ & -0.049  $^{*}$ &  -0.162  $^{**}$ \cr \hline 
    \end{tabular*}
  \end{center}
  Significance levels: $^{*}p<0.01$, $^{**}p<0.001$.
  \caption{{\bf Correlations between word valence and  information measurements.} 
    Correlation coefficients of the valence ($v$), frequency $f$,  self-information $I$, and information content 
    measured for 2-grams $I_2$, 3-grams $I_3$, and 4-grams $I_4$, and with self-information $I'$ 
    measured from the frequencies reported in \cite{Kucera1967, Baayen1993, Sebastian2000}.}
  \label{tab:GGS-2}
\end{table}

Eventually, we also performed a control analysis using alternative
frequency datasets, to account for possible anomalies in the Google
dataset due to its online origin.  We used the word frequencies
estimated from traditional written corpuses, i.e. books, as reported
in the original datasets for English \cite{Kucera1967}, German
\cite{Baayen1993}, and Spanish \cite{Sebastian2000}. Calculating the
self-information from these and relating them to the valences given,
we obtained similar, but slightly lower Pearson's correlation
coefficients $\rho(v, I')$ (see Table \ref{tab:GGS-2}). So, we
conclude that our results are robust across different types of written
communication, for the three languages analyzed.

It is not surprising to find a larger self-information for negative
words, as their probability of appearance is generally lower. The
amount of information carried by a word is also highly dependent on
its context. Among other factors, the context is defined by the word
neighborhood in the sentence. For example, the word ``violent''
contains less information in the sentence ``dangerous murderers are
violent'' than in ``fluffy bunnies are violent'', as the probability
of finding this particular word is larger when talking about murderers
than about bunnies. For this reason we evaluate how the context of a
word impacts its informativeness and valence. The intuition behind
measuring information depending on the context is that the information
content of a word depends primarily on i) the amount of contexts it
can appear and ii) the probability of appearance in each one of these
contexts. Not only the most infrequent, but the most specific and
unexpectable words are the ones that carry the most information. Given
each context $c_i$ where a word $w$ appears, the information content
is defined as
\begin{equation}
-\frac{1}{N} \sum_{i = 1}^{N} {\rm log}({\rm P}(W = w | C=c_i))
\end{equation}
where $N$ is the total frequency of the word in the corpus used for
the estimation. These values were calculated as approximations of the
information content given the words surrounding $w$ up to size $4$.

We analyzed how word valence is related to the information content up
to context size $4$ using the original calculations provided by
Piantadosi et al. \cite{Piantadosi2011}. This estimation is based on
the frequency of sequences of $N$ words, called $N$-grams, from the
Google dataset \cite{Brants2009} for $N\in \{2,3,4\}$. This dataset
contains frequencies for single words and $N$-grams, calculated from
an online corpus of more than a trillion tokens.  The source of this
dataset is the whole Google crawl, which aimed at spanning a large
subset of the web, providing a wide point of view on how humans write
on the Internet. For each size of the context $N$, we have a different
estimation of the information carried by the studied words, with
self-information representing the estimation from a context of size 1.

The left column of Fig. \ref{fig:GGS-2} shows how valence decreases
with the estimation of the information content for each context size.
Each bar represents the same amount of words within a language and has
an area proportional to the rescaled average information content
carried by these words. The color of each bar represents the average
valence of the binned words. The decrease of average valence with
information content is similar for estimations using 2-grams and
3-grams. For the case of 4-grams it also decreases for English and
Spanish, but this trend is not so clear for German. These trends are
properly quantified by Pearson's correlation coefficients between
valence and information content for each context size (Table
\ref{tab:GGS-2}). Each correlation coefficient becomes smaller for
larger sizes of the context, as the information content estimation
includes a larger context but becomes less accurate.


\subsection{Additional analysis of valence, length and self-information}

In order to provide additional support for our results, we tested
different hypotheses impacting the relation between word usage and
valence. First, we calculated Pearson's and Spearman's correlation
coefficients between the absolute value of the valence and the
self-information of a word, $\rho(abs(v), I)$ (see Table
\ref{tab:GGS-1}). We found both correlation coefficients to be around
$0.1$ for German and Spanish, while they are not significant for
English. The dependence between valence and self-information
disappears if we ignore the sign of the valence, which means, indeed,
that the usage frequency of a word is not just related to the overall
emotional intensity, but to the positive or negative emotion expressed
by the word.

\begin{table}[!ht]
  \begin{center}
    \begin{tabular*}{0.8\linewidth}{@{\extracolsep{\fill}}llll}
      \hline
      & English & German & Spanish \cr \hline 
      $\rho(abs(v), I)$ & 0.032  $^{\circ}$ &0.109 $^{***}$ &  0.135 $^{***}$\cr 
      $\rho(l, I)$ &0.378 $^{***}$ & 0.143 $^{***}$ & 0.361 $^{***}$ \cr
      $\rho(v, l)$ & -0.044 $^{\circ}$   &  -0.071 $^{***}$  &  -0.112 $^{***}$  \cr
      $\rho(v, I | l)$ &-0.379 $^{***}$ &-0.319 $^{***}$& -0.399 $^{***}$\cr
      $\rho(l, I | v)$ &0.389 $^{***}$ &0.126 $^{***}$ & 0.357 $^{***}$ \cr \hline
    \end{tabular*}
  \end{center}
  Significance levels: $^{\circ} p<0.3$, $^{*} p<0.1$, $^{**} p<0.01$, $^{***} p<0.001$.
  \caption{{\bf Additional correlations between valence,
      self-information and length.}  Correlation coefficients of the
    valence ($v$), absolute value of the valence ($abs(v)$), and word
    length ($l$) versus self-information ($I$). Partial correlations
    are calculated for both variables ($\rho(v,I|l)$,$\rho(l,I|v)$),
    and correlation between valence and length ($\rho(v,l)$).  }
  \label{tab:GGS-1}
\end{table}

Subsequently, we found that the correlation coefficient between word
length and self-information ($\rho(l, I)$) is positive, showing that
word length increases with self-information. These values of $\rho(l,
I)$ are consistent with previous results \cite{Zipf1935,
  Piantadosi2011}. Pearson's and Spearman's correlation coefficients
between valence and length $\rho(v, l)$ are very low or not
significant. In order to test the combined influence of valence and
length to self-information, we calculated the partial correlation
coefficients $\rho(v, I|l)$ and $\rho(l, I|v)$. The results are shown
in Table \ref{tab:GGS-1}, and are within the 95\% confidence intervals
of the original correlation coefficients $\rho(v, I)$ and $\rho(l,
I)$. This provides support for the existence of an additional
dimension in the communication process closely related to emotional
content rather than communication efficiency. This is consistent with
the known result that word lengths adapt to information content
\cite{Piantadosi2011}, and we discover the independent semantic
feature of valence. Valence is also related to information content but
not to the symbolic representation of the word through its length.

Finally, we explore the sole influence of context by controlling for word
frequency. In Table \ref{tab:GGS-3} we show the partial correlation
coefficients of valence with information content for context sizes
between $2$ and $4$, controlling for self-information. We find that most
of the correlations keep significant and of negative sign, with the
exception of $I_2$ for English. The weaker correlation for context sizes
of $2$ is probably related to two word constructions such as negations,
articles before nouns, or epithets.

\begin{table}[!ht]
  \begin{center}
    \begin{tabular*}{0.8\linewidth}{@{\extracolsep{\fill}}llll}
      \hline
      & English & German & Spanish \cr \hline 
      $\rho(v, I_2 | I)$ & -0.034 $^{\circ}$ & -0.100 $^{***}$& -0.058 $^{*}$\cr
      $\rho(v, I_3 | I)$ & -0.101 $^{**}$ & -0.070 $^{***}$& -0.149  $^{***}$\cr
      $\rho(v, I_4 | I)$ & -0.134 $^{***}$ & -0.020 $^{*}$&  -0.084 $^{**}$\cr \hline
    \end{tabular*}
  \end{center}
  Significance levels: $^{\circ} p<0.3$, $^{*} p<0.1$, $^{**} p<0.01$, $^{***} p<0.001$.
  \caption{ {\bf Partial correlation coefficients between valence and 
      information content.}  Correlation coefficients of the valence 
    ($v$) and information content measured on different context sizes 
    ($I_2$, $I_3$, $I_4$)  controlling for self-information ($I$).}
  \label{tab:GGS-3}
\end{table}

These high-frequent, low-information constructions lead to the
conclusion that $I_2$ does not explain more about the valence than
self-information in English, as short range word interactions change
the valence of the whole particle. This finding supports the
assumption of many lexicon-based unsupervised sentiment analysis
tools, which consider valence modifiers for two-word constructions
\cite{Thelwall2011a, Taboada2011}. On the other hand, the significant
partial correlation coefficients with $I_3$ and $I_4$ suggest that
word information content combines at distances longer than $2$, as
longer word constructions convey more contextual information than
$2$-grams. Knowing the possible contexts of a word up to distance $4$
provides further information about word valence than sole
self-information.

\section{Discussion}

Our analysis provides strong evidence that words with a positive
emotional content are used more often.  This lends support to
Pollyanna hypothesis \cite{Boucher1969}, i.e. positive words are more
often used, for all the three languages studied. Our conclusions are
consistent for, and independent of, different corpuses used to obtain
the word frequencies, i.e. they are shown to hold for traditional
corpuses of formal written text, as well as for the Google dataset and
cannot be attributed as artifacts of Internet communication.

Furthermore, we have pointed out the relation between the emotional
and the informational content of words. Words with negative emotions
are less often used, but because of their rareness they carry more
information, measured in terms of self-information, compared to
positive words.  This relation remains valid even when considering the
context composed of sequences of up to four words
($N$-grams). Controlling for word length, we find that the correlation
between information and valence does not depend on the length, i.e. it
is indeed the usage frequency that matters.

In our analysis, we did not explore the role of syntactic rules and
grammatical classes such as verbs, adjectives, etc. However, previous
studies have shown the existence of a similar bias when studying
adjectives and their negations \cite{Rozin2010}.  The question of how
syntax influences emotional expression is beyond the scope of the
present work. Note that the lexica we use are composed mainly of
nouns, verbs and adjectives, due to their emotional
relevance. Function words such as ``a'' or ``the'' are not considered
to have any emotional content and therefore were excluded from the
original studies. In isolation, these function words do not contain
explicit valence content, but their presence in text can modify the
meaning of neighboring words and thus modify the emotional content of
a sentence as a whole.  Our analysis on partial correlations show
that there is a correlation between the structure of a sentence and
emotional content beyond the simple appearance of individual
words. This result suggests the important role of syntax in the
process of emotional communication. Future studies can extend our
analysis by incorporating valence scores for word sequences,
exploring how syntactical rules represent the link between context and
emotional content.

The findings reported in this paper suggest that the process of
communication between humans, which is known to optimize information
transfer \cite{Piantadosi2011}, also creates a bias towards positive
emotional content.  A possible explanation is the basic impact of
positive emotions on the formation of social links between humans.
Human communication should reinforce such links, which it both shapes
and depends on. Thus, it makes much sense that human languages on
average have strong bias towards positive emotions, as we have shown
(see Figure \ref{fig:GGS-1}).  Negative expressions, on the other
hand, mostly serve a different purpose, namely that of transmitting
highly informative and relevant events. They are less used, but carry
more information.

Our findings are consistent with emotion research in social
psychology. According to \cite{Rime2009}, the expression of positive
emotions increases the level of communication and strengthens social
links.  This would lead to stronger pro-social behaviour and
cooperation, giving evolutionary advantage to societies whose
communication shows a positive bias. As a consequence, positive
sentences would become more frequent and even advance to a social norm
(cf. ``Have a nice day''), but they would provide less information
when expressed. Our analysis provides insights on the asymmetry of
evaluative processes, as frequent positive expression is consistent
with the concept of \textit{positivity offset} introduced in
\cite{Miller1961} and recently reviewed in \cite{Norman2011}. In
addition, Miller's \textit{negativity bias} (stronger influence of
proximal negative stimuli) found in experiments provides an
explanation for the higher information content of negative
expression. When writing, people could have a tendency to avoid
certain negative topics and bring up positive ones just because it
feels better to talk about nice things. That would lower the frequency
of negative words and lower the amount of information carried by
positive expression, as negative expression would be necessary to
transmit information about urgent threats and dangerous events.

Eventually, we emphasize that the positive emotional ``charge'' of
human communication has a further impact on the quantitative analysis
of communication on the Internet, for example in chatrooms, forums,
blogs, and other online communities. Our analysis provides an
estimation of the emotional baseline of human written expression, and
automatic tools and further analyses will need to take this into
account. In addition, this relation between information content and
word valence might be useful to detect anomalies in human emotional
expression. Fake texts supposed to be written by humans could be
detected, as they might not be able to reproduce this spontaneous
balance between information content and positive expression.

\section{Acknowledgments}
 This research has received funding from the European Community's
  Seventh Framework Programme FP7-ICT-2008-3 under grant agreement no
  231323 (CYBEREMOTIONS).

\bibliographystyle{unsrt}

\end{document}